\documentclass[10pt,twocolumn,letterpaper]{article}
\usepackage[accsupp]{axessibility} 
\usepackage[pagenumbers]{cvpr} %

\usepackage{graphicx}
\usepackage{amsmath}
\usepackage{amssymb}
\usepackage{booktabs}
\usepackage{multirow}
\usepackage{soul}
\usepackage{pifont}%

\usepackage{algorithm}
\usepackage{algorithmic}
\usepackage[pagebackref,breaklinks,colorlinks]{hyperref}

\usepackage[capitalize]{cleveref}
\crefname{section}{Sec.}{Secs.}
\Crefname{section}{Section}{Sections}
\Crefname{table}{Table}{Tables}
\crefname{table}{Tab.}{Tabs.}
\crefname{algorithm}{Alg.}{Algs.}
\crefname{algorithm}{Algorithm}{Algorithms}

\newcommand{\cmark}{\ding{51}}%
\newcommand{\xmark}{\ding{55}}%

\def\confName{CVPR}
\def\confYear{2022}

\begin{document}
\title{Unified Transformer Tracker for Object Tracking}

\author{Fan Ma$^{1, 3}\thanks{This work was done during Fan's internship at Meta AI.}$ 
~~Mike Zheng Shou$^{2}$ 
~~Linchao Zhu$^{1}$ \\ 
~~Haoqi Fan$^{3}$ 
~~Yilei Xu$^{3}$
 ~~Yi Yang$^{4}$ 
 ~~Zhicheng Yan$^{3}$\thanks{Correspondence author.} \\
$^{1}$ReLER Lab, AAII, University of Technology Sydney 
~~$^{2}$National University of Singapore \\
~~$^{3}$Meta AI
~~$^{4}$Zhejiang University
}

\maketitle
\begin{abstract}
As an important area in computer vision, object tracking has formed two separate communities that respectively study Single Object Tracking (SOT) and Multiple Object Tracking (MOT).
However, current methods in one tracking scenario are not easily adapted to the other due to the divergent training datasets and tracking objects of both tasks.
Although UniTrack~\cite{wang2021different} demonstrates that a shared appearance model with multiple heads can be used to tackle individual tracking tasks, it fails to exploit the large-scale tracking datasets for training and performs poorly on the single object tracking.
In this work, we present the Unified Transformer Tracker (UTT) to address tracking problems in different scenarios with one paradigm. 
A track transformer is developed in our UTT to track the target in both SOT and MOT where the correlation between the target feature and the tracking frame feature is exploited to localize the target. 
We demonstrate that both SOT and MOT tasks can be solved within this framework, and the model can be simultaneously end-to-end trained by alternatively optimizing the SOT and MOT objectives on the datasets of individual tasks.
Extensive experiments are conducted on several benchmarks with a unified model trained on both SOT and MOT datasets.

\end{abstract}

\begin{figure}[t]
  \centering
   \includegraphics[width=\linewidth]{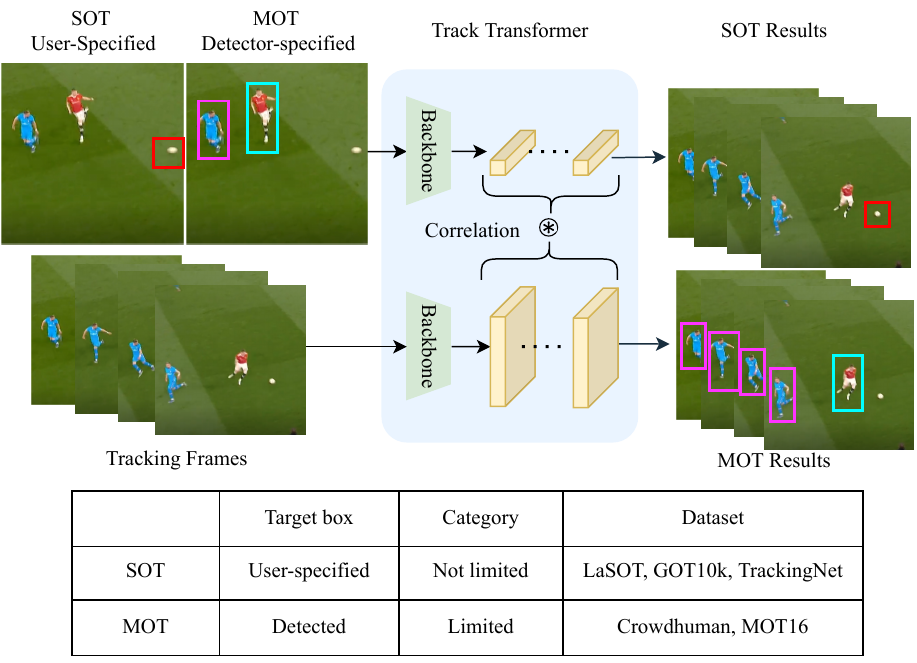}
   \caption{\textbf{Unified transformer tracker for both single object tracking (SOT) and multiple object tracking (MOT) task}. The target box in SOT is specified in the first frame while all boxes in reference frames are from the detection model in MOT. We use one tracking model to predict target localization in tracking frames for both tasks.}
   \label{fig:intro}
\end{figure}

\section{Introduction}
\label{sec:intro}
Visual object tracking is one of the fundamental computer vision tasks with numerous applications~\cite{li2019siamrpn++, siamban, sun2021transtrack, zhang2020fairmot}. 
Unlike clearly defined object classification and detection problems \cite{he2016deep,wu2020dynamic,he2017mask,yang2022colar}, object tracking is considered in different scenarios and can be categorized into two main paradigms 1) Single Object Tracking (\textbf{SOT}) is to track an annotated target from any object category in the first frame throughout a video ~\cite{li2019siamrpn++, danelljan2020probabilistic};  
 2) Multiple Object Tracking (\textbf{MOT}) aims at estimating bounding boxes and identities of objects in videos where categories of targets are known and objects could appear or disappear~\cite{zhang2020fairmot, zeng2021motr}. 
 Current methods in the tracking community solve individual tasks separately by training models on the individual datasets for either SOT or MOT.

Siamese architecture is widely applied in SOT where various designs focus on improving the discriminative representation of objects~\cite{li2019siamrpn++,siamban, danelljan2020probabilistic, Yang2020SiamAttSA}.
For MOT, tracking by detection is the most popular paradigm and achieves highest tracking performance on several benchmarks~\cite{sadeghian2017tracking, bergmann2019tracking, zhang2020fairmot}.
This paradigm is not applicable for SOT as the model would fail to detect objects of unseen categories in SOT.
Some MOT methods~\cite{leal2016learning, zhu2018online} use the Siamese tracker in SOT~\cite{bertinetto2016fully} to predict the location of the targets in tracking frames and fuse the predicted boxes with the detection boxes to enhance the detection results.
However, these methods are not competitive to the tracking-by-detection methods in MOT.
Albeit the Siamese trackers have been applied in both tracking tasks, none of these works are able to address two tracking tasks with a unified paradigm.
In practice, a unified tracking system is significant in many fields.
For the AR/VR applications, tracking specific or unseen instances like personal cups is related to SOT while perceiving the environment of general classes like people is related to MOT. It is expensive and inefficient to maintain two separate tracking systems. The unified tracking system, which can easily switch tracking mode by demands, becomes more essential in real world deployment.

UniTrack~\cite{wang2021different} firstly attempts to address SOT and MOT concurrently by sharing the backbone and fusing multiple tracking heads. 
However, It fails to exploit large scale tracking datasets for training due to the head design and the divergent training datasets in different tasks.
As shown in the \cref{fig:intro}, the training data in SOT and MOT are from various sources. Datasets in SOT only provide annotations of a single target in one video while dense object annotations are available in MOT datasets although the object categories are fixed. 
The tracking capacity of UniTrack~\cite{wang2021different} is thus limited, and the model would fail to track objects in complex scenarios.

In this paper, we introduce a Unified Transformer Tracker (UTT) as shown in \cref{fig:intro} for solving both tracking tasks. For the tracking object in the reference frame, which is specified in SOT or detected in MOT, we provide a small feature map proposal in the tracking frame based on the previous localization.
The target feature then correlates with the feature map proposal to update target representation and output the target localization. 
This enables our UTT to track objects in both SOT and MOT with the same design.
The updated target features are further correlated with the new search feature proposal, which is cropped based on the produced target localization.  This process is repeated several times to refine the localization of tracking targets.
To enable exploiting training samples from both tasks, we alternatively train the network with datasets in each task. Experiments demonstrate that our tracker can be well learned on both tasks.
Our main contributions are summarized as follows:

\begin{itemize}
  \item We propose a novel Unified Transformer Tracker (UTT) for both single and multiple object tracking. To the best of our knowledge, this is the first work that the tracking model is end-to-end trained on both tasks. 
  \item We develop a novel and effective track transformer to localize targets via the correlation between target feature and the tracking frame feature. The target features are well encoded through our transformer design.
  \item To verify the unified object tracking capability, we evaluate our UTT on both SOT and MOT benchmarks. Our proposed method achieves comparable performance to the state-of-the-art algorithms not only on LaSOT~\cite{fan2019lasot}, TrackingNet~\cite{muller2018trackingnet}, and GOT-10k~\cite{huang2019got} in the SOT setting, but also on MOT16~\cite{Milan2016MOT16AB} in the MOT setting.
\end{itemize}

\begin{figure*}[!ht]
\vspace{-0.5cm}

  \centering
   \includegraphics[width=1.0\linewidth]{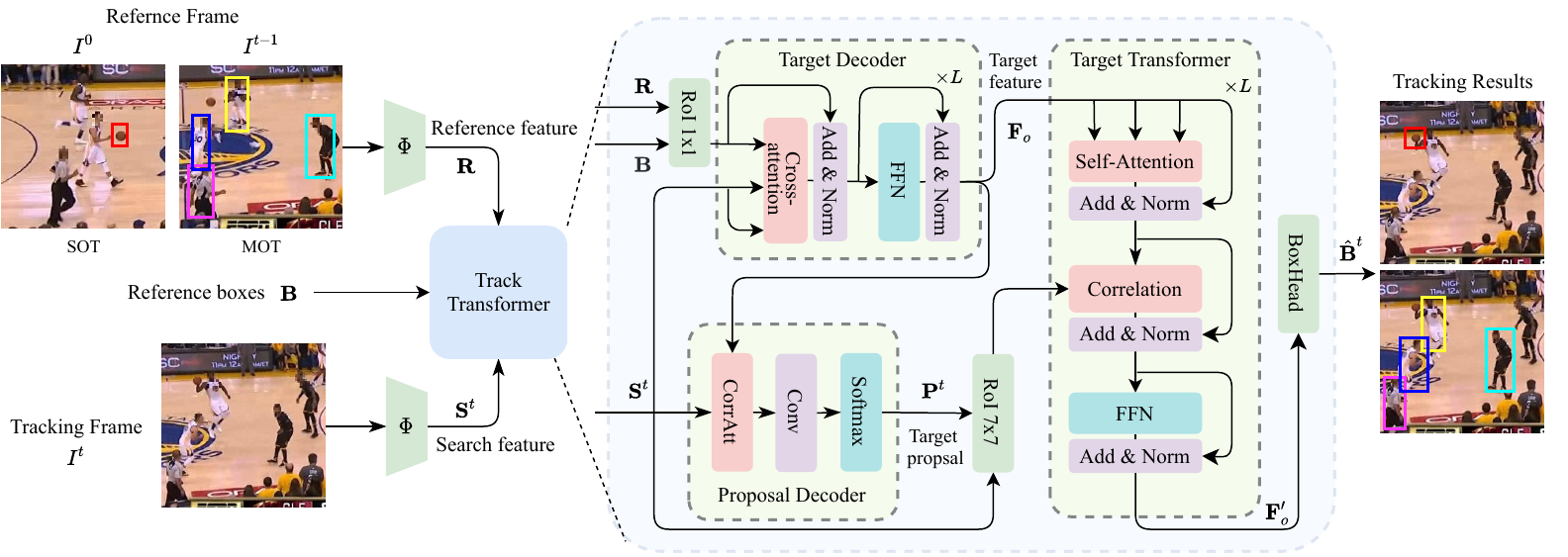}

   \caption{\textbf{Framework of our unified transformer tracker (UTT)}. We first use the backbone $\Phi$ to extract frame features. There are three inputs to the track transformer, including frame features of the reference and tracking frames, and reference boxes in the reference frame. The goal of track transformer is to predict the target localization in the tracking frame. The target decoder in the track transformer is first used to extract target features, and the proposal decoder is to produce candidate search areas in the tracking frame. Both target features and search features are fed to the target transformer to predict the target localization. Notations are detailed in the \cref{subsec:tracktrans}. 
   }
   \vspace{-0.5cm}
   \label{fig:framework}
\end{figure*}

\section{Related Work}
\label{sec:relate}

\subsection{Single Object Tracking}
The goal of single object tracking is to estimate the location of the object in a video sequence, while only the initial position is provided and the object category is unknown.
It has been undergone astonishing progress in recent years, with the development of a variety of approaches, ranging from early correlation filter based methods ~\cite{bolme2010visual,henriques2014high,danelljan2014accurate}, and recent Siamese network based trackers~\cite{bertinetto2016fully,zhu2018distractor,bhat2019learning}. 
The pioneering work by Bolme~\emph{et al.}~\cite{bolme2010visual} proposes a minimum output sum of square error (MOSSE) filter.
 Siamese trackers has been used in various SOT methods~\cite{bertinetto2016fully,wang2018learning, li2019siamrpn++, siamban}, which learn a general similarity map by cross-correlation between two image regions. 
Recently, transformer~\cite{NIPS2017_3f5ee243} has also been applied in the single object tracking community to improve the tracking performance via learning discriminative target representation~\cite{Chen_2021_CVPR, Wang_2021_CVPR, Yu_2021_ICCV}.  
Most SOT methods crop the targets and tracking frames in the reference image for extracting target representation.
However, cropping every target and tracking frame becomes inefficient when multiple objects are specified in the tracking scenario.
Differently, our tracker extracts target representation on the high-level feature map and constricts the search area by cropping feature maps. Rather than using the common cross attention in the transformer, we adopt the correlation attention between the target feature and search feature to update target representation for tracking.

\subsection{Multiple Object Tracking}
Multi-object trackers focus on tracking an unknown number of objects from a fixed set of categories~\cite{dave2020tao}, which is often built on top of the tracking-by-detection paradigm via linking the detections across successive frames~\cite{zhang2008global,jiang2007linear}.
Bea~\emph{et al.}~\cite{bae2017confidence} introduce the Siamese architecture for learning a dissimilarity metric between a pair of objects.
FairMOT~\cite{zhang2020fairmot} learn the object detection task and appearance embedding task from a shared backbone to improve the association accuracy. 
CenterTrack~\cite{Zhou2020TrackingOA} proposes to detect objects with center points and predict their offsets to the previous frame for tracking. 
Recently, several methods~\cite{sun2021transtrack,zeng2021motr} adopt the transformer-based detectors~\cite{detr_eccv2020, Zhu2021DeformableDD} to propagate boxes between frames. 
In these methods, the encoder with self attention is first adopted to enhance feature representation, and the decoder is followed with cross attention where the learnable query features are replaced with detected object features in previous frames.
Our tracker integrates the encoder and decoder into one object transformer where the self attention is applied on the multiple target features and correlation attention is then adopted to update target features with the search features.

\subsection{Unified Tracking}
UniTrack~\cite{wang2021different} tackles both SOT and MOT problem via a shared backbone. For each task, a unique parameter-free task head is designed to cope with the tracking scenario. ImageNet pretrained features are directly used for all the tracking tasks, and the training recipe on all tracking tasks is not disclosed. It is thus still unknown whether the multiple tracking heads can be trained together on different tracking datasets. 
Besides, the tracking performance is not competitive to that of current state-of-the-art methods especially on the single object tracking task.
In this paper, we propose a unified transformer tracker where different tasks share the same tracking head. It enables our tracker to be trained on both SOT and MOT datasets, and tracking performance on SOT is substantially improved.

\section{Method}
\label{sec:method}

\subsection{Overview}
Existing tracking tasks can be  categorized into single object tracking (SOT) and multiple-object tracking (MOT). We review their definitions below.

\noindent{\textbf{SOT}}.
For a video sequence containing $T$ frames, the tracking object is specified in the first frame $I^{0}$.  Let $\mathbf{B}^{0}=\{(x_1,y_1,x_2,y_2)\}$ denote the object coordinates in the first frame,  the goal of SOT is to localize the target $\{\hat{\mathbf{B}}^{t}\}_{t=1}^{T}$ in all $T$ frames.
In SOT, the tracking targets could be an object from any category, and the objects in testing data could belong to the unseen categories. The tracking performance is directly evaluated by comparing the predicted localization $\{\hat{\mathbf{B}}^{t}\}_{t=1}^{T}$  with the groundtruth localization $\{\mathbf{B}^{t}\}_{t=1}^{T}$.

\noindent{\textbf{MOT}}. 
 Objects from a fixed set of categories are detected and tracked in MOT. 
Suppose we have $N$ detected objects $\mathbf{B}_{d}^{t-1} = \{b_{i}^{t-1}\}_{i=1}^{N}$ in the previous tracking frame $I^{t-1}$ and $M$ detected boxes $\mathbf{B}_{d}^{t} = \{b_{i}^{t}\}_{i=1}^{M}$  in the tracking frame $I^{t}$.
The number of tracking objects is not fixed among frames as the new objects may appear and old objects may disappear in videos. 
The core challenge in MOT is to associate the previous $N$ objects with current $M$ objects, where the same object in frames should be assigned with a unique ID.

In our tracker, we predict the target boxes in the tracking frame $\hat{\mathbf{B}}^{t} \in \mathcal{R}^{N\times 4}$ for previous $N$ objects, and then match them with current detected boxes $\mathbf{B}_{d}^{t}$. 
The performance of accurately tracked boxes $\{\hat{\mathbf{B}}^{t}\}_{t=1}^{T}$ in each frame is key to the overall MOT performance.

Taking a unified perspective, we treat the initial frame in SOT and the previous frame in MOT as the reference frame $I$, and the initial box in SOT and detected boxes in MOT as the reference boxes $\mathbf{B}$.
The goal in our tracker is then to predict accurate target localization for tracking frame $I^{t}.$
We propose a Unified Transformer Tracker as shown in ~\cref{fig:framework} to address both SOT and MOT tasks. 
For the reference frame $I$, we first extract frame feature using the backbone as $\mathbf{R} = \Phi(I) \in \mathcal{R}^{H\times W\times C}$ where $H$ and $W$ is the height and width of the feature map. The frame feature can be extracted from different layers, here we only consider one layer for simplicity. The current tracking frame feature $\mathbf{S}^{t}$ is also extracted in the same way. Then we feed these frame features with target boxes $\mathbf{B}$ to the track transformer for predicting target localization.

\subsection{Track Transformer}
\label{subsec:tracktrans}
The track transformer is to produce the target coordinates in the tracking frame given reference boxes $\mathbf{B}$, the reference feature $\mathbf{R}$ and the tracking frame feature $\mathbf{S}^{t}$. The tracking process can be expressed by 
\begin{equation}
    \hat{\mathbf{B}}^{t} = \mathcal{T}_{\theta}(\mathbf{S}^{t}, \mathbf{R}, \mathbf{B}),
\end{equation}
where $\hat{\mathbf{B}}^{t} \in \mathcal{R}^{N\times4}$ denotes the predicted localization for tracking targets. 
$\mathcal{T}_{\theta}$ denotes the track transformer parameterized by $\theta$. 

To localize targets in the tracking frame, we first extract the target feature using the \textit{target decoder} in \cref{fig:framework}. 
For each tracking target, we then adopt the \textit{proposal decoder} to generate candidate search area that may contain targets in the tracking frame.
The target feature and proposal are further processed in the \textit{target transformer} to predict the localization in the tracking frame.

\subsubsection{Target Decoder}
An intuitive way of forming target features is to apply the RoIAlign~\cite{he2017mask} on the reference feature map, calculated by
\begin{equation}
    \mathbf{F} = \text{RoIAlign}(\mathbf{R}, \mathbf{B}) \in \mathcal{R}^{N\times C}.
\end{equation}

However, this would neglect spatial background information, which is critical in object tracking especially when different tracking targets are similar in appearance.
To embed more context information,  we adopt Multi-head Cross Attention (MCA) to interacts the target feature with the frame feature. The target feature is then update through
\begin{equation}
    \mathbf{F}_{c} = \text{Norm}(\mathbf{F} + \text{MCA}(\mathbf{F}, \mathbf{S}^{t}, \mathbf{S}^{t})),   
    \label{eq:cross_attention}
\end{equation}
where the target feature $\mathbf{F}$  is the query and the frame feature $\mathbf{S}^{t}$ represents the key and value in the MCA module. Norm denotes the instance normalization. The frame feature is flattened along with spatial dimensions so that there are $HW$ values and keys. 

In addition, we use a standard FFN module, which contains a fully connected feed-forward network that consists of two linear transformations with a ReLU in between. The output target feature is calculated by
\begin{equation}
    \mathbf{F}_{o} = \text{Norm}(\mathbf{F}_{c} + \text{FFN}(\mathbf{F}_{c})).
    \label{eq:ffn}
\end{equation}

With the target decoder, we embed more context information into the target representation through the cross attention mechanism.

\subsubsection{Proposal Decoder}
We track every target by correlating the target feature with the tracking frame feature. 
However, correlating each target feature with the whole frame feature would incur huge memory and computational cost when the track frame is of high resolution and multiple objects are required to be tracked, 
We introduce the \textit{target proposal} to crop a unique search area for each object to reduce the search area in the tracking frame. 
A naive way of setting target proposal would be choosing it to be the tracking result of the last frame $\mathbf{P}^{t} = \hat{\mathbf{B}}^{t-1}$.
However, the tracker would fail to localize the target if the previous tracking box $\hat{\mathbf{B}}^{t-1}$ is full of background and no tracking target is contained. 
The issue becomes more severe when the specified tracking target disappears from the video and appears at a new location later on.
For instance, the tracking target in SOT could be lost if there are no targets in previous predicted localization. It is critical to generate more accurate target proposal in our track transformer.

To this end, we correlate the target feature with tracking frame feature for providing an effective target proposal in SOT. Specifically, we obtain the heat maps for targets via  
\begin{equation}
    \mathbf{H} = \text{Softmax}(\text{Conv}(\text{CorrAtt}(\mathbf{F}_{o}, \mathbf{S}^{t}))) \in \mathcal{R}^{N \times HW \times 2},
     \label{eq:softmax}
\end{equation}
where \textit{Conv} contains several convolutional operations with ReLU activations. The \textit{Softmax} operation is executed on the flattened the spatial dimension. The \textit{CorrAtt} operation correlates the target feature with the tracking frame feature via
\begin{equation}
    \text{CorrAtt}(\mathbf{F}_{o}, \mathbf{S}^{t}) =  (\mathbf{F}_{o} * \mathbf{S}^{t}) \odot \mathbf{S}^{t},
    \label{eq:dyn}
\end{equation}
where $*$ is the convolution operation and $\mathbf{F}_{o}$ is the $1\times1$ filter with one output channel. $\odot$ is the broadcast element-wise product. 

The heat map $\mathbf{H}$ indicates the probability distribution of top-left and bottom-right of targets.
We follow the process in \cite{Yan_2021_ICCV} to generate the target proposal by
\begin{equation}
    \mathbf{P}^{t} = \sum_{i=0}^{W}\sum_{j=0}^{H}(i*\mathbf{H}^{0}_{i,j},j*\mathbf{H}^{0}_{i,j},i*\mathbf{H}^{1}_{i,j},j*\mathbf{H}^{1}_{i,j}),
    \label{eq:proposal}
\end{equation}
where $i$ and $j$ denote the x and y coordinate, respectively. $\mathbf{H}^{0}_{i,j}$ and $\mathbf{H}^{1}_{i,j}$ denote the probability of top-left and right-bottom of the target at the localization $(i, j)$, respectively.

\subsubsection{Target Transformer}
We first crop a search feature map for each object in the tracking frame given the proposals $\mathbf{P}^{t}$ by
\begin{equation}
    \mathbf{S}_{RoI}^{t} = \text{RoIAlign}(\mathbf{S}^{t}, \mathbf{P}^{t}) \in \mathcal{R}^{N \times K \times K \times C},
\end{equation}
where $K$ is the size of the output feature map. In this way, we have a candidate search feature for each tracking target.
We employ the attention mechanism to update target features via
\begin{equation}
    \mathbf{F}_{a} = \text{Norm}(\mathbf{F}_{o} + \text{MSA}(\mathbf{F}_{o})),  
    \label{eq:msa}
\end{equation}
where MSA denotes the multi-head self-attention.
We also use a sine function to generate spatial positional encoding following \cite{Zhu2021DeformableDD}.

To capture the target information in the search area,  we connect the target feature with search feature map through
\begin{equation}
    \mathbf{F}_{d} = \text{Norm}(\mathbf{F}_{a} + \text{Correlation}(\mathbf{F}_{a}, \mathbf{S}^{t}_{RoI}))
\end{equation}
where Correlation updates the target feature through 
\begin{equation}
    \text{Correlation}(\mathbf{F}_{a}, \mathbf{S}^{t}_{RoI}) =  \text{FC}(\text{CorrAtt}(\mathbf{F}_{a}. \mathbf{S}^{t}_{RoI})),
    \label{eq:interact}
\end{equation}
where \textit{FC} denotes the fully connected layer with the input channel $K^{2}C$ and output channel $C$.  
The \textit{CorrAtt} is the same as the operation in \cref{eq:dyn}.
In addition, a \textit{FFN} module in \cref{eq:ffn} is applied to produce target features $\mathbf{F}_{o}'$ as shown in \cref{fig:framework}.
The output feature is then used to predict the target localization. Instead of directly producing coordinates, the box head outputs the delta bias to the proposal boxes for generating target localization. 

We repeat the target transformer for $L$ times to localize tracking targets. 
In the $i^{th} (i>1)$ target transformer, the target proposal is updated by the previous predicted localization to provide new candidate search area. Similarly, the target feature is also updated to correlate with the search feature. 
All the intermediate outputs are stored in $\{\hat{\mathbf{B}}^{t}_{i}\}_{i=1}^{L}$.

\noindent\textbf{Differences from the original transformer.}
In the tracking community, all the previous transformer designs \cite{Chen_2021_CVPR, Wang_2021_CVPR, Yu_2021_ICCV, Yan_2021_ICCV, sun2021transtrack} use encoder-decoder pipeline to enhance the target representation. The target feature is treated as the query and the whole tracking frame feature is the key and value in the cross-attention operation. It becomes inefficient when the multiple specified objects are required to be tracked and the video is of high resolution. Suppose we have $N$ objects to track and the size of the tracking frame feature map is $H\times W$ and the feature dimension is $C$. The complexity of the cross attention for all tracking objects will be $\mathcal{O}(HWNC)$. In contrast, we crop the fixed size $K\times K (K < min(H, W))$ search area from the frame feature. The complexity in our tracker is $\mathcal{O}(K^{2}NC)$, which are more efficient.
Moreover, the cross attention is replaced with our correlation attention, which is more effective for tracking objects in various scenarios.
In general, our tracker is not only able to be end-to-end trained on both SOT and MOT tasks, but also achieves higher performance when testing on SOT and MOT datasets.

\begin{table*}[!ht]
\vspace{-0.5cm}

\centering

\resizebox{1.5\columnwidth}{!}{

\begin{tabular}{l|l|cc|cc|ccc}
\toprule
\multirow{2}{*}{Type} & \multirow{2}{*}{Method}    & \multicolumn{2}{c}{LaSOT~\cite{fan2019lasot}} & \multicolumn{2}{|c}{TrackingNet~\cite{muller2018trackingnet}}  & \multicolumn{3}{|c}{GOT10K\cite{huang2019got}} \\ 
& &  Success   & Precision    & Success   & Precision       & AO & SR$_{0.5}$   & SR$_{0.75}$ \\
\midrule

\multirow{19}{*}{SOT-Only} &SiamFC~\cite{bertinetto2016fully}  &33.6 &33.9 &57.1 &66.3  &34.8 &35.3 &9.8 \\
&MDNet~\cite{nam2016learning}   &39.7 &37.3 &60.6 &56.5  &29.9 &30.3 &9.9 \\
&ECO~\cite{danelljan2017eco}   &32.4 &30.1 &55.4 &49.2  &31.6 &30.9 &11.1 \\
&VITAL~\cite{Song2018VITALVT}   &39.0 &36.0 &- &-   &35.0 &36.0 &9.0 \\
&GradNet~\cite{Li2019GradNetGN}    &36.5 &35.1 &- &-  &- &- &- \\
&SiamDW~\cite{Zhang2019DeeperAW}   &38.4 &35.6 &- &-  &41.6 &47.5 &14.4 \\
&SiamRPN++~\cite{li2019siamrpn++}   &49.6 &49.1 &73.3 &69.4  &51.7 &61.6 &32.5 \\
&ATOM~\cite{danelljan2019atom}   &51.5 &50.5 &70.3 &64.8 &55.6 &63.4 &40.2\\
&DiMP~\cite{bhat2019learning}  &56.9 &56.7 &74.0 &68.7  &61.1 &71.7 &49.2\\
&SiamFC++~\cite{Xu2020SiamFCTR}  &54.3 &54.7 &75.4 &70.5  &59.5 &69.5 &47.9 \\
&D3S~\cite{Lukei2020D3SA}  &- &- &72.8 &66.4  &59.7 &67.6 &46.2 \\
&MAMLTrack~\cite{Wang2020TrackingBI}  &52.3 &53.1 &75.7 &72.5  &- &- &- \\
&SiamAttn~\cite{Yang2020SiamAttSA}  &56.0 &- &75.2 &- &- &- &- \\
&SiamCAR~\cite{Guo2020SiamCARSF}  &50.7 &51.0 &- &-  &56.9 &67.0 &41.5 \\
&SiamBAN~\cite{siamban}  &51.4 &52.1 &- &-  &- &- &- \\
&KYS~\cite{bhat2020know}  &55.4 &55.8 &74.0 &68.8  &63.6 &75.1 &51.5\\
&Ocean~\cite{Zhang2020OceanOA}  &52.6 &52.6 &70.3 &68.8  &59.2 &69.5 &46.5\\
&TrDiMP~\cite{Wang_2021_CVPR} & 63.9 & 61.4 &78.4 & 73.1  & 68.8 & 80.5 & 59.7 \\
&TransT~\cite{Chen_2021_CVPR} & 64.9 & 69.0 & 81.4 &80.3  & 72.3 &82.4 &68.2 \\ 
\midrule
\multirow{2}{*}{Unified} & UniTrack~\cite{wang2021different}  & 35.1  & 32.6  & 59.1  & 51.2  & -  & -  & -  \\
& UTT (ours)    & 64.6  & 67.2  &79.7 & 77.0 & 67.2 & 76.3	& 60.5  \\
\bottomrule
\end{tabular}

}
\caption{
\textbf{Comparisons on TrackingNet, LaSOT, and GOT-10k}. All results are reported from original manuscripts. 
Our tracker outperforms the UniTrack by a large margin on the SOT datasets.
}
\vspace{-0.5cm}
\label{tab:sot_comp}
\end{table*}

\subsection{Training}
To train our tracker, we calculate the loss between the predicted boxes and ground truth boxes. 
For MOT, we do not use the proposal decoder to generate initial candidate search areas for training. Instead, we add a Guassian noise to the ground truth boxes $\mathbf{B}^{t}$ for producing the initial proposals during training $\mathbf{P}^{t}$. 
This is more similar to the testing phrase where detected boxes in previous frames are used as the proposals in the current tracking frame.
Moreover, the model is unable to be trained if we produce the proposal with the proposal decoder module for all objects. 
This is due to the high resolution of the feature map and  a large quantity of objects in MOT.
Let $\mathbf{B}^{t} \in \mathcal{R}^{N\times 4}$ be the ground truth boxes in $t^{th}$ frame, the loss function in MOT can be written as
\begin{equation}
    \mathcal{L}_{MOT}^{box} = \sum_{i=1}^{L} \lambda_{G}\mathcal{L}_{GIoU} (\hat{\mathbf{B}}^{t}_{i}, \mathbf{B}^{t}) + \lambda_{1} \mathcal{L}_{1}(\hat{\mathbf{B}}_{i}^{t}, \mathbf{B}^{t}),
    \label{eq:mot}
\end{equation}
where $\mathcal{L}_{GIoU}$ and $\mathcal{L}_{1}$ denote the generalized IoU loss~\cite{rezatofighi2019generalized} and 1-norm
loss, respectively. $\lambda_{G}$ and $\lambda_{1}$ are the hyper-parameters to balance the box loss.

For single object tracking, the target proposal is initialized by the proposal decoder in \cref{fig:framework}. To produce a robust proposal for the single object, we also consider the box loss between the initial proposal and the ground truth $\mathbf{B}^{t}$.
The loss function for the SOT is then expressed by
\begin{equation}
    \mathcal{L}_{SOT}^{box} = \sum_{i=0}^{L} \lambda_{G}\mathcal{L}_{GIoU} (\hat{\mathbf{B}}^{t}_{i}, \mathbf{B}^{t}) + \lambda_{1} \mathcal{L}_{1}(\hat{\mathbf{B}}_{i}^{t}, \mathbf{B}^{t}),
    \label{eq:sot}
\end{equation}
where the initial prediction $\hat{\mathbf{B}}_{i}^{0} = \mathbf{P}^{t}$ is calculated in the target decoder via \cref{eq:proposal}.

For single and multiple object tracking, we train the model with Eq.~\eqref{eq:sot} and Eq.~\eqref{eq:mot}, respectively. 
Note that the training datasets are different in the two tasks. SOT datasets only provide a unique target annotation in one video where the dense boxes with classes are given in the MOT datasets. 
Moreover, object detection is necessary in the MOT as new objects should be also tracked.
To this end, we simply add a detection head in our UTT for the detection task.
In this paper, we implement the deformable DETR~\cite{Zhu2021DeformableDD} with a set-based Hungarian loss~\cite{detr_eccv2020} to optimize the detection head. More details of the training paradigm with the detection head are given in the appendix.

To enable a single model for both SOT and MOT tasks, we alternatively train the network with SOT and MOT datasets at each iteration.
Specifically, we implement two data loaders for each task. 
At each iteration, two frames in a video from one data loader are sampled and batched as inputs to our tracker for optimization. 
For the SOT iteration,  we use the target decoder to improve target representation and use the proposal decoder to produce initial target proposals in tracking frames. 
For the MOT iteration, all annotated objects are used for optimizing the detection head and objects in both frames are used for optimizing the track transformer with \cref{eq:mot}.
We provide the detailed optimization procedure in the appendix.

\section{Experiment}
\label{sec:exp}

\subsection{Implementation Details}

\noindent\textbf{Training}.
We train our Unified transformer tracker on the training splits
of COCO~\cite{lin2014microsoft}, TrackingNet~\cite{muller2018trackingnet}, LaSOT~\cite{fan2019lasot}, GOT10k~\cite{huang2019got}, MOT16~\cite{Milan2016MOT16AB}, and CrowdHuman~\cite{Shao2018CrowdHumanAB} datasets.
ResNet~\cite{resnet} is used as the backbone with ImageNet-pretrained weights.
The whole model is trained on 8 NVIDIA DGX A100 GPUs for 800K iterations with AdamW optimizer~\cite{Loshchilov2019DecoupledWD}.
We use the cosine scheduler for decaying the learning rate from 1e-4 to 1e-5 with the weight decay 5e-3. We have two data loaders for SOT and MOT training, and the two tasks are alternatively trained in the unified training mode. 
For the SOT setting, we set the batch size as 64, and images are cropped to 352x352 with common SOT data augmentions~\cite{danelljan2020probabilistic}. The SOT data loader contains both image and video datasets. The COCO~\cite{leal2016learning} is applied with different transformations for generating image pairs during training.
For the rest of video datasets, we randomly sample images with frame interval less than 200. 
For the MOT setting, we set the batch size as 16, and images are randomly resized following the data augmentations in MOT methods~\cite{sun2021transtrack}. We also add a deformable DETR~\cite{Zhu2021DeformableDD} head in our model for detecting objects.

\noindent\textbf{Online Tracking}.
We evaluate our model on several datasets in both SOT and MOT tasks. For the single object tracking, we test our model on the testing splits of LaSOT~\cite{fan2019lasot}, TrackingNet~\cite{muller2018trackingnet}, and GOT10K~\cite{huang2019got} datasets. For the multiple object tracking, we validate our model on MOT16~\cite{Milan2016MOT16AB} in two modes, using private or public object detection.

\begin{table*}[!t]
\vspace{-0.5cm}

\centering

\resizebox{1.6\columnwidth}{!}{

\begin{tabular}{l|l|cccccccc}
\toprule
Type & Method  & MOTA$\uparrow$ & IDF1$\uparrow$ & MOTP$\uparrow$ & MT$\uparrow$ & ML$\downarrow$ & FP$\downarrow$ & FN$\downarrow$ & IDs$\downarrow$ \\
\midrule
\multirow{7}{*}{MOT-Only} &TAP~\cite{Zhou2018OnlineMT}  &64.8 &73.5 &78.7 &292 &164 &12980 &50635 & 571 \\
 &CNNMTT~\cite{Mahmoudi2018MultitargetTU}  &65.2 &62.2 &78.4 &246 &162 &6578 &55896 &946 \\
 &POI~\cite{Yu2016POIMO}  &66.1 &65.1 &79.5 &258 &158 & 5061 &55914 &3093 \\
 &TubeTK\_POI~\cite{Pang2020TubeTKAT}  &66.9 &62.2 &78.5 &296 &122 &11544 &47502 &1236\\
 &CTrackerV1~\cite{Peng2020ChainedTrackerCP}  &67.6 &57.2 &78.4 &250 &175 &8934 &48305 &1897 \\
 &FairMOT~\cite{zhang2020fairmot}  &74.9 & 72.8 &- &447 &159 &- &- & 1074 \\
 &TransTrack~\cite{sun2021transtrack} & 74.5 &63.9	& 80.6 & 468	& 113	& 28323	& 112137	& 3663 \\
 \midrule
 \multirow{3}{*}{Unified} &UniTrack~\cite{wang2021different}  & 74.7 & 71.8 &- &- &- &- &- &683 \\
 &UTT-DFDETR         &  69.2     & 53.9      & 80.1  & 417 & 39   & 22917  & 29050 & 4145   \\
 &UTT-FairMOT          &  74.2     & 63.4      & 81.4  & 322 & 140   & 7863  & 37147 & 2002   \\
\bottomrule

\end{tabular}

}

\caption{\textbf{Comparisons on MOT Datasets}.  $\uparrow$ means the higher the better and $\downarrow$ means the lower the better. UTT-DFDETR denotes that the deformable DETR~\cite{Zhu2021DeformableDD} detection head is contained in our tracker while the UTT-FairMOT indicates that we use the detected objects from FairMOT~\cite{zhang2020fairmot}. Without using any person re-identification features, our tracker achieves highest MOTP among all methods.}
\label{tab:mot}
\vspace{-0.3cm}
\end{table*}

\subsection{Evaluations in SOT}
\label{sec:exp_comp}

We compare our UTT with 20 state-of-the-art SOT methods on three datasets as shown in \cref{tab:sot_comp}. Moreover, our track can run at 25 FPS on the SOT datasets.

\noindent\textbf{LaSOT.}
LaSOT~\cite{fan2019lasot} contains 1,400 sequences with 1,120 for training and 280 for testing.
We compare different trackers on LaSOT test set using Success and Precision metric. The results are reported in Tab.~\ref{tab:sot_comp}.
Our UTT outperforms most of SOT trackers (\emph{i.e.,} SiamRPN++~\cite{li2019siamrpn++} and KYS~\cite{bhat2020know}) on all datasets, and is competitive to the best tracker TransT~\cite{Chen_2021_CVPR}. 
Moreover, our proposed tracker outperforms UniTrack~\cite{wang2021different} by almost 30 points on the Success metric.

\noindent\textbf{TrackingNet.}
TrackingNet~\cite{muller2018trackingnet} consists of 30k video sequences, and 511 videos are contained in the test set.
As shown in \cref{tab:sot_comp}, compared with the UniTrack~\cite{wang2021different}, our UTT obtains 25.8$\%$ and 33.5$\%$ higher Success and Precision.
Moreover, our method is competitive with the best task-specific SOT algorithm TransT~\cite{Chen_2021_CVPR}, which crops both target and tracking frame and is unable to cope with multiple object tracking.  

\noindent\textbf{GOT10K.}
GOT10K~\cite{huang2019got} collects over 10k video segments and contains 180 testing sequences. We conduct the experiments and submit the tracking results to the official online server for evaluation. 
The  detailed results of AO, SR$_{0.5}$ and SR$_{0.75}$ are reported in \cref{tab:sot_comp}.
Our tracker outperforms most of Siamese trackers, including SiamRPN++~\cite{li2019siamrpn++}, SiamFC++~\cite{Xu2020SiamFCTR}, and SiamAttn~\cite{Guo2020SiamCARSF}. Our unified tracker achieves second higher SR$_{0.75}$ among all methods.

\subsection{Evaluations in MOT}
\label{sec:exp_comp_mot}
We also assess our method on the MOT16~\cite{Milan2016MOT16AB} for evaluation on the MOT task. The MOT16~\cite{Milan2016MOT16AB} benchmark consists of a train set
and a test set, each with 7 sequences and pedestrians annotated with full-body bounding boxes.
Different aspects of MOT are evaluated by a number of individual metrics \cite{bernardin2008evaluating}, and the results on MOT16 are reported in \cref{tab:mot}. 
We report UTT with different detection results on this dataset. 
First, we use the deformable DETR~\cite{Zhu2021DeformableDD} as our detection head and train the detection head together with our track transformer (denoted by UTT-DFDETR). Our tracker runs at 8 FPS on this dataset. UTT-DFDETR outperforms recent CTracker~\cite{Peng2020ChainedTrackerCP} on the MOTA metric, which also uses ResNet50 as the backbone and manages to match same objects by using tracking localization.  
Following the UniTrack~\cite{wang2021different}, our tracker also uses the detection results from FairMOT~\cite{zhang2020fairmot}.  Instead of using person re-identification features to match detection results in FairMOT~\cite{zhang2020fairmot} or using Kalman filter to estimate target movement in UniTrack~\cite{wang2021different}, our method directly predicts target localization given target features. 
Our tracker achieves best MOTP score (81.4) among all methods, and attains slightly lower MOTA score compared to FairMOT~\cite{zhang2020fairmot}.

\subsection{Ablative Studies}
\label{sec:ablation}
To verify the effectiveness of our designed transformer structure,  we use the ResNet-18 as the backbone and train the model for 100K iterations.
We validate our method on the LaSOT test set in the SOT setting and on the half training set in MOT following~\cite{Zhou2020TrackingOA}.

\begin{table}[]
\vspace{-0.2cm}
\centering

\resizebox{0.9\columnwidth}{!}{

\begin{tabular}{cc|ccc}
\toprule
  Target decoder & Proposal decoder  & Success & Precision & OP75\\ 
\midrule
\xmark &\xmark & 48.1 & 44.7  & 38.3 \\
\cmark &\xmark & 49.8  & 48.5 & 43.9  \\
\xmark &  \cmark & 58.6 & 59.4  & 53.9 \\
\cmark &\cmark & 59.2 & 59.8  & 54.3 \\

\bottomrule
\end{tabular}
}

\caption{\textbf{Ablation study on the target and proposal decoder}. We report the Success, Precision, and OP75 on the LaSOT~\cite{fan2019lasot} for comparisons. The tracking performance is largely improved by using target and proposal decoders.}
\label{tab:proposal}
\vspace{-0.3cm}
\end{table}

\subsubsection{Target- and Proposal Decoder}

To validate the target and proposal decoder on the tracking performance, we conduct the experiments on LaSOT and report the Success, Precision and OP75 metrics in \cref{tab:proposal}.
For the baseline model, the target feature is simply extracted with RoIAlign and the target proposal is the tracking result of the last frame.
The tracking performance is largely improved when we add the proposal decoder.
This is due to the fact that the tracking localization in previous frames could provide false proposals, containing no objects, and the localization would not be rectified in the tracking sequence.
We further add the target decoder as proposed in~\cref{fig:framework} to enhance the target feature representation by interacting the target feature with the tracking frame feature, the tracking performance is also improved by 0.6 on the Success metric.

\begin{table}[!t]
\centering

\resizebox{0.9\columnwidth}{!}{

\begin{tabular}{l|cc|cc}
\toprule
 & \multicolumn{2}{c}{LaSOT~\cite{fan2019lasot} }  & \multicolumn{2}{|c}{MOT16~\cite{Milan2016MOT16AB}}  \\
 & Success & Precision & MOTA & IDF1 \\
\midrule
MCA  & 57.1 & 56.9 & 64.4 & 57.7 \\
CorrAtt   & 57.6 &57.9 & 64.9 & 61.4 \\
CorrAtt + MSA    & 58.9 &59.4 & 65.4 & 59.9 \\
\bottomrule

\end{tabular}
}
\caption{\textbf{Comparing different correlation designs in the target transformer}. 
}
\vspace{-0.5cm}
\label{tab:interact}
\end{table}

\subsubsection{Target Transformer}

\noindent\textbf{Correlation.}
To show the superiority of our correlation layer in the target transformer in \cref{eq:interact}, we design a tracker using the multi-head cross attention (MCA)~\cite{Chen_2021_CVPR, Wang_2021_CVPR}. Specifically, we replace the correlation module and self-attention module in \cref{fig:framework}  with the MCA module, where the target features are treated as query and cropped search features are used as key and value. We also remove the multi-head self-attention (MSA) to testify our designed module. The comparison results are shown in \cref{tab:interact}. 
Compared to the UTT-MCA tracker, our method with the correlation improves the Success by 1.8\% on the SOT dataset, and MOTA by 1.0\% on the MOT dataset.

\begin{figure}
\vspace{-0.3cm}
  \centering
  \begin{subfigure}{0.48\linewidth}
    \includegraphics[scale=0.3]{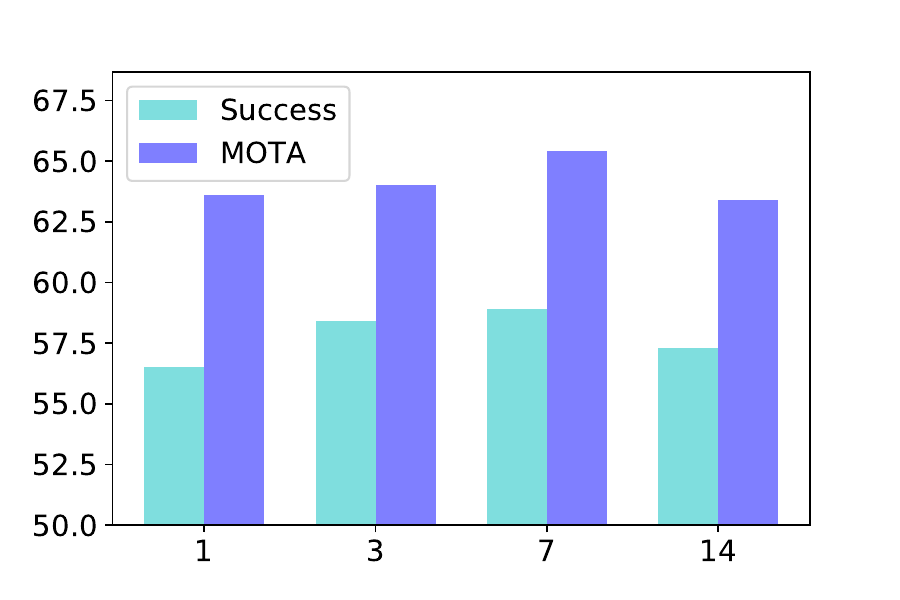}
    \caption{Search feature size $K$.}
    \label{fig:pool}
  \end{subfigure}
  \begin{subfigure}{0.48\linewidth}
    \includegraphics[scale=0.3]{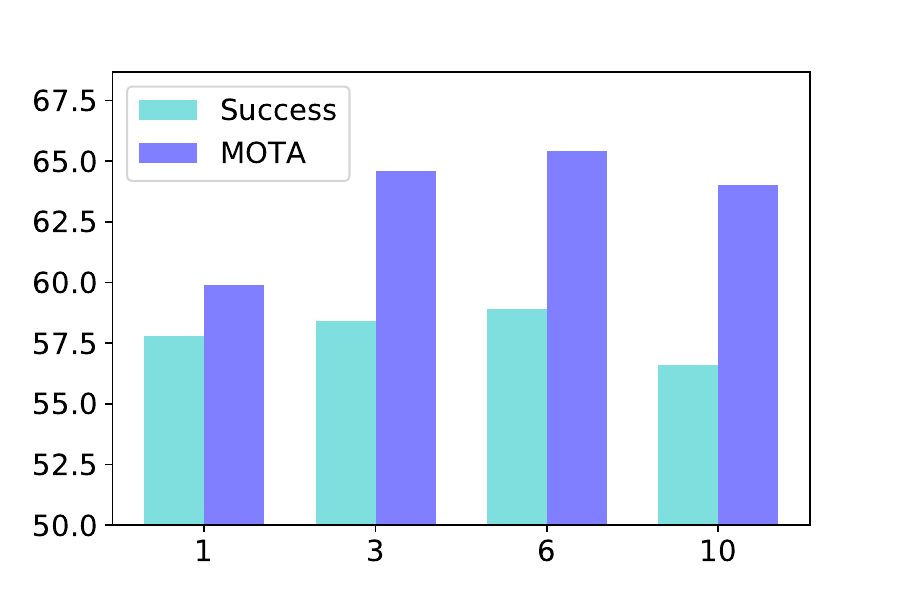}
    \caption{Target transformer iteration $L$.}
    \label{fig:iteration}
  \end{subfigure}
  \caption{\textbf{Comparing different hyperparameter choices}. Success is the performance metric on LaSOT~\cite{fan2019lasot} and MOTA is the metric on MOT16~\cite{Milan2016MOT16AB}. 
  }
 \vspace{-0.5cm}
  \label{fig:short}
\end{figure}

\noindent\textbf{Search size.}
For each tracking target, we crop the search feature given the target proposal.
The cropped feature interacts with the target feature for updating the localization. We thus validate the model with search features of different sizes in~\cref{fig:pool}.
Highest tracking performance on both tasks is achieved when the pool size is set to 7.

\noindent\textbf{Iteration.}
We testify the tracking performance with different number $L$ of target transformer in~\cref{fig:iteration}.
It shows that the model with more iterations indeed achieves higher tracking performance. However, with more iterations, the training and inference speed will be slower as more parameters are contained and computational costs are higher.

\begin{table}[]
\vspace{-0.2cm}
\centering

\resizebox{0.8\columnwidth}{!}{
\begin{tabular}{l|cc|cc}
\toprule
\multirow{2}{*}{Training}   & \multicolumn{2}{c}{LaSOT\cite{fan2019lasot}}  & \multicolumn{2}{|c}{MOT16\cite{Milan2016MOT16AB}} \\ 
& Success   & Precision   & MOTA  & IDF1  \\
\midrule
SOT-Only & 59.2 & 59.8 &- &-  \\
MOT-Only & 12.5 & 5.1 & 64.3  & 62.1  \\ 
Unified & 58.9 & 59.4 & 65.4 & 59.9 \\
\bottomrule
\end{tabular}
}

\caption{\textbf{Comparisons of different training modes}. Our model trained with both SOT and MOT datasets (Unified training) can be used for both tracking scenarios.
}
\vspace{-0.2cm}
\label{tab:trainingmode}
\end{table}

\subsubsection{Training Recipe}
To investigate the tracking performance when trained in different settings, we conduct experiments with three different training modes in~\cref{tab:trainingmode}. The model is trained on COCO~\cite{lin2014microsoft}, LaSOT~\cite{fan2019lasot}, GOT10K~\cite{huang2019got}, and TrackingNet~\cite{muller2018trackingnet} in SOT, and is trained on CrowdHuman~\cite{Shao2018CrowdHumanAB} and MOT16\cite{Milan2016MOT16AB} in MOT. 
All these datasets are used in the unified training mode.
As shown in \cref{tab:trainingmode}, the model trained on SOT is not able to track objects in MOT as no detection head is trained, while the model trained with MOT datasets performs poorly on the SOT datasets as only pedestrian annotations are provided.
Compared to the MOT-Only, our unified model trained on both SOT and MOT datasets achieves higher MOTA and lower IDF1.
The decreased IDF1 means that more IDs are assigned to a single object when dismissed or occluded (higher IDSW). Higher MOTA indicates that the false negative and false positive tracking objects decreased, demonstrating that our unified model provides more accurate tracking results.

\begin{figure}[]
  \centering
   \includegraphics[width=1.0\linewidth]{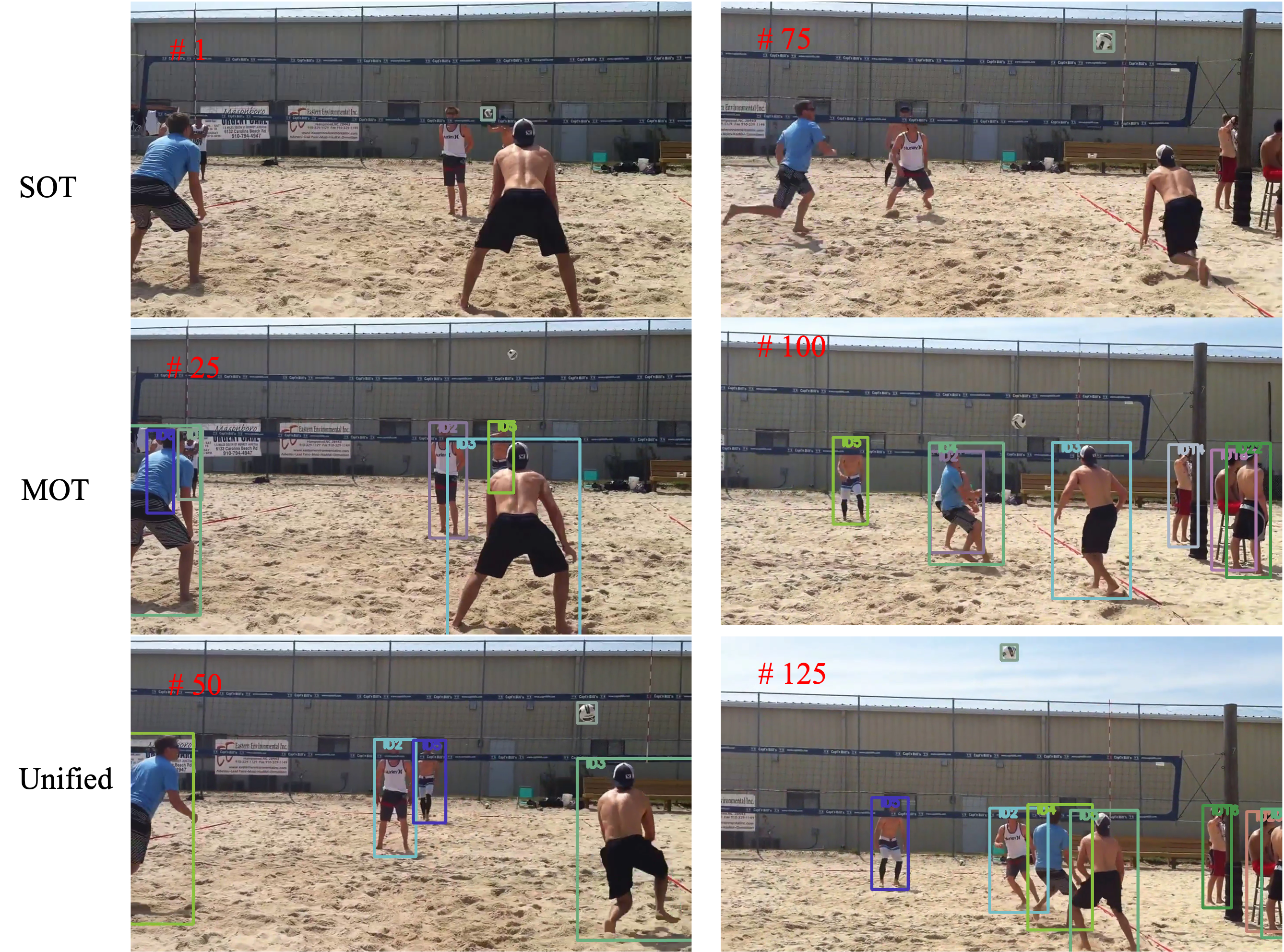}

   \caption{\textbf{Visualization of our results on the challenging sequences in different tracking modes}. Best viewed with zooming in. In the unified tracking mode, the specified volleyball is tracked together with multiple persons. 
   }
  \vspace{-0.5cm}
   \label{fig:visualize}
\end{figure}

\subsubsection{Visualization}
We visualize the tracking results with different tracking modes in \cref{fig:visualize}.
The unified tracking mode indicates that our model does SOT and MOT concurrently, where the box of the volleyball is specified and boxes of persons are detected by the deformable DETR head~\cite{Zhu2021DeformableDD}.
We run the sequence three times with different tracking modes and the results in different timestamps are displayed. In the MOT and Unified tracking mode, our models assign the unique identification for each tracked person with the same color.  From the figure, we observe that the person occluded by others can be also tracked well. It shows that both SOT and MOT tasks can be done concurrently during testing by our single model. The video demo of this sequence with different tracking modes is attached in the supplementary material.

\section{Conclusion}
Can a single model address different tracking tasks?
We propose a Unified Transformer Tracker (UTT) to solve this problem, where the model can be end-to-end trained with datasets in both SOT and MOT tasks.
A novel track transformer is introduced in this work to handle different tracking scenarios.
From the experiments, we can draw the conclusion that one model is enough to address both SOT and MOT tasks.
We believe this will encourage the community to develop more unified tracking algorithms
that are applied into more wild tracking scenarios. 

\noindent\textbf{Acknowledgements}
Mike Shou is solely supported by the National Research Foundation, Singapore under its NRFF Award NRF-NRFF13-2021-0008.
Linchao Zhu is partially supported by Australian Research Council Discovery Projects DP200100938.

{\small
\bibliographystyle{ieee_fullname}
\bibliography{egbib}
}

\clearpage

\appendix

\begin{figure}[!ht]
  \centering
  \includegraphics[width=1.0\linewidth]{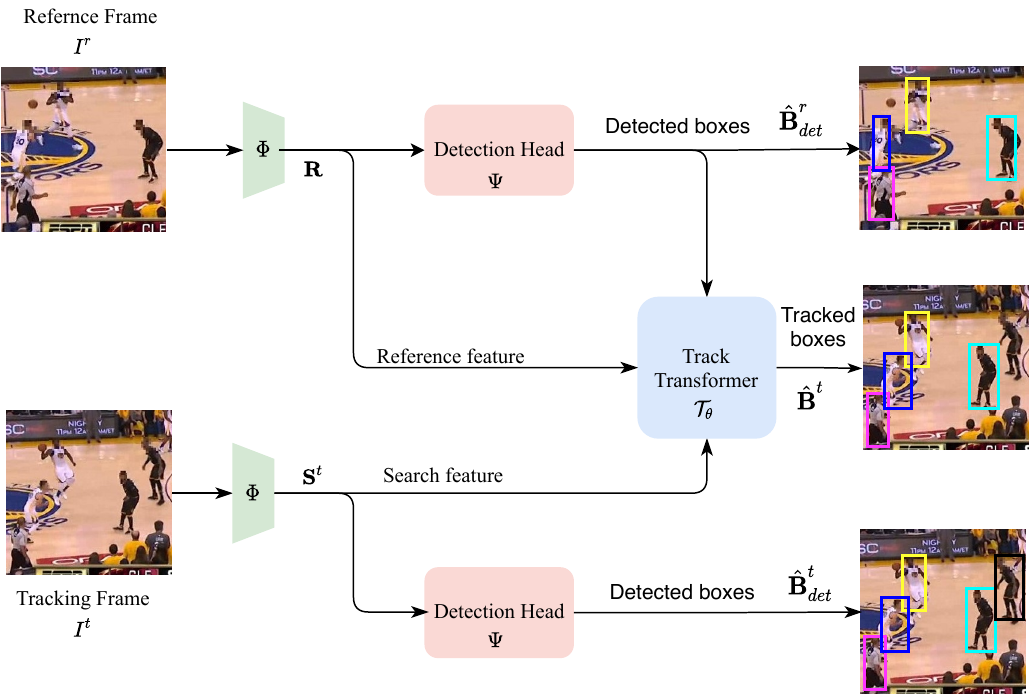}

  \caption{UTT with the detection head.
  }
  \label{fig:framework_det}
\end{figure}

\section{MOT Detection \& Tracking}
In the multiple object tracking, objects should be detected and tracked. In ~\cref{fig:framework}, we only include the track transformer to predict target localization given previous target coordinates.
In this part, we display our Unified Transformer Tracker with the detection head in ~\cref{fig:framework_det}. In this work, we use the deformable DETR~\cite{Zhu2021DeformableDD} as the detection head $\Psi$. For the MOT training, the backbone $\Phi$, detection head $\Psi$, and the track transformer $\mathcal{T}_{\theta}$ are trained together. 

During tracking, we first detect all the objects $\hat{\mathbf{B}}^{r}_{det}$ in the reference frame $I^{t}$.
The detected boxes $\hat{\mathbf{B}}^{r}_{det}$ are then used as target proposals for the tracking frame.
We feed the tracking frame feature, target proposals $\hat{\mathbf{B}}^{r}_{det}$, and the reference feature to the track transformer and produce the tracked boxes $\hat{\mathbf{B}}^{t}$. Moreover, the detection head also produces the object detection results $\hat{\mathbf{B}}^{t}_{det}$ in the tracking frame.
By calculating the IoU between the tracked boxes $\hat{\mathbf{B}}^{t}$ and the detected boxes $\hat{\mathbf{B}}^{t}_{det}$, we can match the same objects in both previous reference and current tracking frames. 
The IoU threshold is set to 0.9 for matching the same objects.
For the unmatched tracked boxes, we mark them as the lost objects and track them in the latter frames.
For the unmatched detected boxes, we assign new identifications to them as they are the new objects appearing in the video.
In this way, we track all objects in the MOT and assign the same objects with the same identifications.

\begin{algorithm}[!t]
	
	\begin{algorithmic}[1]
		\caption{Unified Training of UTT on both SOT and MOT}
	    \label{alg:mix_training}
		
		\STATE{\bfseries Input:} $\mathcal{D}_{sot}$, $\mathcal{D}_{mot}$, backbone $\Phi$, detection head $\Psi$ and tracker transformer $\mathcal{T}_{\theta}$
		\FOR {\textit{iter} $\leftarrow 0$ to \textit{max\_iter}}
		\STATE Updating model with SOT datasets via \cref{alg:sot_iteration} \\
		$Alg_{sot}(\mathcal{D}_{sot}, \Phi, \mathcal{T}_{\theta})$
		\STATE Updating model with MOT datasets via \cref{alg:mot_iteration} \\
		$Alg_{mot}(\mathcal{D}_{mot}, \Phi, \Psi, \mathcal{T}_{\theta})$
		\ENDFOR
		
	\end{algorithmic}
\end{algorithm}

\begin{algorithm}[!t]
	
	\begin{algorithmic}[1]
		\caption{SOT Iteration $Alg_{sot}$}
	    \label{alg:sot_iteration}
		
		\STATE{\bfseries Input:} $\mathcal{D}_{sot}$, backbone $\Phi$ and tracker transformer $\mathcal{T}_{\theta}$
		\STATE Sampling ($I^{r}$, $I^{t}$, $\mathbf{B}^{r}$, $\mathbf{B}^{t}$) from $\mathcal{D}_{sot}$
        \STATE Predicting target localization in reference frames $I^{r}$:\\
        $\hat{\mathbf{B}}^{r} = \mathcal{T}_{\theta} (\Phi(I)^{r}, \Phi(I)^{r}, \mathbf{B}^{r}) = \{\hat{\mathbf{B}}^{r}_{i}\}_{i=0}^{L}$
        \STATE Predicting target localization in tracking frames $I^{t}$:\\
        $\hat{\mathbf{B}}^{t} = \mathcal{T}_{\theta} (\Phi(I)^{t}, \Phi(I)^{r}, \mathbf{B}^{r}) = \{\hat{\mathbf{B}}^{t}_{i}\}_{i=0}^{L}$
        \STATE Calculating SOT loss via ~\cref{eq:sot}:\\
        \resizebox{1.0\linewidth}{!}{%
        $\mathcal{L}_{SOT}^{box} = \sum\limits_{j\in\{t, r\}}\sum\limits_{i=0}^{L} \lambda_{G}\mathcal{L}_{IoU} (\hat{\mathbf{B}}^{j}_{i}, \mathbf{B}^{j}) + \lambda_{1} \mathcal{L}_{1}(\hat{\mathbf{B}}_{i}^{j}, \mathbf{B}^{j})$}
        \STATE Updating $\Phi$ and $\mathcal{T}_{\theta}$
		
	\end{algorithmic}
\end{algorithm}

\begin{algorithm}[!t]
	
	\begin{algorithmic}[1]
		\caption{MOT Iteration $Alg_{mot}$}
    	\label{alg:mot_iteration}
		
		\STATE{\bfseries Input:} $\mathcal{D}_{mot}$, backbone $\Phi$, detection head $\Psi$ and tracker transformer $\mathcal{T}_{\theta}$
		\STATE Sampling ($I^{r}$, $I^{t}$, $\mathbf{B}^{r}$, $\mathbf{B}^{t}$) from $\mathcal{D}_{sot}$
        \STATE Predicting object detection in reference frames $I^{r}$:\\
        $\hat{\mathbf{B}}^{r}_{det} = \Psi (\Phi(I)^{r}) $
        \STATE Calculating the detection loss: \\
        $\mathcal{L}_{MOT}^{det} = SetCriterion(\hat{\mathbf{B}}^{r}_{det} , \mathbf{B}^{r})$
        \STATE Predicting target localization in tracking frames $I^{t}$:\\
        $\hat{\mathbf{B}}^{t} = \mathcal{T}_{\theta} (\Phi(I)^{t}, \Phi(I)^{r}, \mathbf{B}^{r}) = \{\hat{\mathbf{B}}^{t}_{i}\}_{i=1}^{L}$
        \STATE Calculating MOT tracking loss via ~\cref{eq:mot}:\\
        $\mathcal{L}_{MOT}^{box} = \sum\limits_{i=1}^{L} \lambda_{G}\mathcal{L}_{IoU} (\hat{\mathbf{B}}^{t}_{i}, \mathbf{B}^{t}) + \lambda_{1} \mathcal{L}_{1}(\hat{\mathbf{B}}_{i}^{t}, \mathbf{B}^{t})$
        \STATE Calculating the MOT loss: \\
        $\mathcal{L}_{MOT} = \mathcal{L}_{MOT}^{box} + \mathcal{L}_{MOT}^{det}$
        \STATE Updating $\Phi$, $\Psi$, and $\mathcal{T}_{\theta}$
		
	\end{algorithmic}
\end{algorithm}

\begin{table*}[!t]
\centering
\resizebox{\linewidth}{!}{%
\begin{tabular}{clllllllllllllllllll}
\toprule
\textbf{seq}        & \textbf{MOTA}$\uparrow$ & \textbf{MOTP}$\uparrow$ & \textbf{IDF1}$\uparrow$ & \textbf{IDP}$\uparrow$ & \textbf{IDR}$\uparrow$ & \textbf{TP}$\uparrow$ & \textbf{FP}$\downarrow$ & \textbf{FN}$\downarrow$ & \textbf{Rcll}$\uparrow$ & \textbf{Prcn}$\uparrow$ & \textbf{MTR}$\uparrow$ & \textbf{PTR} & \textbf{MLR}$\downarrow$ & \textbf{MT}$\uparrow$ & \textbf{PT}$\downarrow$ & \textbf{ML}$\downarrow$ & \textbf{IDSW}$\downarrow$ & \textbf{FAR}$\downarrow$ & \textbf{FM}$\downarrow$ \\ 
\midrule
\textbf{MOT16-01}             & 57.36         & 81.60         & 43.17         & 52.24        & 36.78        & 4116        & 386         & 2279        & 64.36         & 91.43         & 43.48        & 39.13        & 17.39        & 10          & 9           & 4           & 62            & 0.86         & 73          \\
\textbf{MOT16-03}             & 89.18         & 82.45         & 73.16         & 74.87        & 71.53        & 96764       & 3134        & 7792        & 92.55         & 96.86         & 87.84        & 12.16        & 0.00         & 130         & 18          & 0           & 385           & 2.09         & 434         \\
\textbf{MOT16-06}             & 61.93         & 80.48         & 59.35         & 71.76        & 50.60        & 7736        & 400         & 3802        & 67.05         & 95.08         & 33.48        & 42.08        & 24.43        & 74          & 93          & 54          & 190           & 0.34         & 191         \\
\textbf{MOT16-07}             & 63.96         & 80.25         & 47.60         & 53.87        & 42.64        & 11780       & 1139        & 4542        & 72.17         & 91.18         & 40.74        & 55.56        & 3.70         & 22          & 30          & 2           & 201           & 2.28         & 251         \\
\textbf{MOT16-08}             & 44.32         & 81.16         & 36.72         & 53.21        & 28.03        & 8231        & 585         & 8506        & 49.18         & 93.36         & 30.16        & 46.03        & 23.81        & 19          & 29          & 15          & 228           & 0.94         & 236         \\
\textbf{MOT16-12}             & 58.30         & 80.14         & 62.05         & 73.67        & 53.60        & 5479        & 556         & 2816        & 66.05         & 90.79         & 33.72        & 44.19        & 22.09        & 29          & 38          & 19          & 87            & 0.62         & 100         \\
\textbf{MOT16-14}             & 46.32         & 74.68         & 45.88         & 56.23        & 38.74        & 11073       & 1663        & 7410        & 59.91         & 86.94         & 23.17        & 48.78        & 28.05        & 38          & 80          & 46          & 849           & 2.22         & 442         \\
\textbf{OVERALL}              & 74.22         & 81.39         & 63.36         & 69.42        & 58.27        & 145179      & 7863        & 37147       & 79.63         & 94.86         & 42.42        & 39.13        & 18.45        & 322         & 297         & 140         & 2002          & 1.33         & 1727   \\
\bottomrule

\end{tabular}
}
\caption{Detailed results on MOT16 test set with UTT FairMOT. All results are reported from online evaluation sever. The detected boxes are from FairMOT~\cite{zhang2020fairmot}.
}

\label{tab:mot16_fairmot}
\end{table*}

\begin{table*}[!t]
\centering
\resizebox{1.0\linewidth}{!}{%
\begin{tabular}{clllllllllllllllllll}
\toprule
\textbf{seq}        & \textbf{MOTA}$\uparrow$ & \textbf{MOTP}$\uparrow$ & \textbf{IDF1}$\uparrow$ & \textbf{IDP}$\uparrow$ & \textbf{IDR}$\uparrow$ & \textbf{TP}$\uparrow$ & \textbf{FP}$\downarrow$ & \textbf{FN}$\downarrow$ & \textbf{Rcll}$\uparrow$ & \textbf{Prcn}$\uparrow$ & \textbf{MTR}$\uparrow$ & \textbf{PTR} & \textbf{MLR}$\downarrow$ & \textbf{MT}$\uparrow$ & \textbf{PT}$\downarrow$ & \textbf{ML}$\downarrow$ & \textbf{IDSW}$\downarrow$ & \textbf{FAR}$\downarrow$ & \textbf{FM}$\downarrow$ \\ 
\midrule
\textbf{MOT16-01} & 47.27          & 78.03         & 41.49         & 45.27        & 38.30        & 4265        & 1145        & 2130        & 66.69         & 78.84         & 39.13        & 47.83        & 13.04        & 9           & 11          & 3           & 97            & 2.54         & 137         \\
\textbf{MOT16-03} & 87.00 & 80.99         & 62.98         & 65.36        & 60.76        & 94373       & 2819        & 10183       & 90.26         & 97.10         & 84.46        & 14.19        & 1.35         & 125         & 21          & 2           & 588           & 1.88         & 997         \\
\textbf{MOT16-06} & 55.23 & 78.63         & 51.69         & 52.03        & 51.36        & 9124        & 2266        & 2414        & 79.08         & 80.11         & 50.68        & 44.34        & 4.98         & 112         & 98          & 11          & 485           & 1.90         & 326         \\
\textbf{MOT16-07} & 58.49 & 80.17         & 42.59         & 43.07        & 42.13        & 12904       & 3062        & 3418        & 79.06         & 80.82         & 50.00        & 50.00        & 0.00         & 27          & 27          & 0           & 295           & 6.12         & 368         \\
\textbf{MOT16-08} & 31.76 & 79.15         & 35.45         & 32.33        & 39.22        & 13042       & 7258        & 3695        & 77.92         & 64.25         & 57.14        & 42.86        & 0.00         & 36          & 27          & 0           & 468           & 11.61        & 482         \\
\textbf{MOT16-12} & 44.70 & 81.89         & 59.52         & 56.30        & 63.13        & 6550        & 2752        & 1745        & 78.96         & 70.41         & 52.33        & 45.35        & 2.33         & 45          & 39          & 2           & 90            & 3.06         & 199         \\
\textbf{MOT16-14} & 39.39 & 76.16         & 34.82         & 36.75        & 33.07        & 13018       & 3615        & 5465        & 70.43         & 78.27         & 38.41        & 48.78        & 12.80        & 63          & 80          & 21          & 2122          & 4.82         & 565         \\
\textbf{OVERALL}  & 69.22 & 80.17         & 53.94         & 54.88        & 53.03        & 153276      & 22917       & 29050       & 84.07         & 86.99         & 54.94        & 39.92        & 5.14  & 417         & 303         & 39          & 4145          & 3.87         & 3074       \\
\bottomrule
\end{tabular}
}
\caption{Detailed results on MOT16 test set with UTT$-$DFDETR. All results are reported from online evaluation sever. The detected boxes are produced using our deformable detection head.}
\label{tab:mot16_dfdetr}
\end{table*}

\section{Unified Training}
To train our Unified transformer tracker on both SOT and MOT tasks, we create two data loaders $\mathcal{D}_{sot}$ and $\mathcal{D}_{mot}$. The detailed procedure to update models with both datasets is described in \cref{alg:mix_training}.

Specifically, we use $\mathcal{D}_{sot}$ and $\mathcal{D}_{mot}$ to update the model in each iteration. In SOT, we only calculate the tracking box loss to update the backbone $\Phi$ and the track transformer $\mathcal{T}_{\theta}$. In the SOT iteration, we use the target decoder to extract target features, and use the proposal decoder to produce candidate search areas for the target transformer. The produced proposal $\hat{\mathbf{B}}_{0}^{t}$ is thus used to calculate the loss in \cref{eq:sot}. For the MOT iteration, the detection head $\Psi$ is also adopted to predict object detection in frames. In this work, we employ the Deformable DETR~\cite{Zhu2021DeformableDD} as the detection head.
The detection head predicts both box localization and classification results. For simplicity, we only present the box notation in \cref{alg:mot_iteration}. The set criterion loss in \cite{detr_eccv2020} is used to optimize the detection part. 
Also we calculate the tracking loss with \cref{eq:mot}. Instead of using target proposals, the target proposals in the MOT iteration are produced by adding Guassian noise to the ground truth boxes. To ensure that the target is included in the proposal, we set the minimum IoU as 0.1 between the ground truth proposal and the noise proposal.
The backbone $\Phi$, detection head $\Psi$, and the track transformer $\mathcal{T}_{\theta}$ are then updated with both detection and tracking losses.

\section{Detailed Results on MOT16}
The tracking results are submitted to the online server for evaluation. We report the detailed scores on every testing video in \cref{tab:mot16_fairmot} and \cref{tab:mot16_dfdetr}. The UTT-DFDETR denots the model trained with deformable detection head, and all the objects are dected using the detection head.
The UTT-FairMOT uses the detected boxes from FairMOT~\cite{zhang2020fairmot}, only track transformer is used in this model to track detected object in previous reference frames.

\section{Limitations}
We propose a unified transformer tracker to tracking objects in different scenarios. Ideally, the tracker is required to track objects over 30 FPS. Currently, our tracker is not able to do the online tracking on the MOT task. Another limitation is that our tracker in the present paper is only trained with pedestrian in the MOT task. However, objects of various categories should be considered (such the vehicles) although this part is more related to the detection head. We mainly focus on the unified transformer tracker in the current manuscript.

\section{Code}
We will release our code to the public later.
Our implementation is mainly based on Detectron2\footnote{\url{https://github.com/facebookresearch/detectron2}} and Pytracking\footnote{\url{https://github.com/visionml/pytracking}}. Our implementation supports distributed training and testing. Previous methods (especially the Siamese pipelines) can be easily reproduced with our code.

\end{document}